\documentclass[journal=eds]{CUP-JNL-DTM}%

\usepackage{graphicx}
\usepackage{multicol,multirow}
\usepackage{amsmath,amssymb,amsfonts}
\usepackage{mathrsfs}
\usepackage{amsthm}
\usepackage{rotating}
\usepackage{appendix}
\usepackage{ifpdf}
\usepackage[T1]{fontenc}
\usepackage{newtxtext}
\usepackage{newtxmath}
\usepackage{textcomp}
\usepackage{xcolor}
\usepackage{booktabs}
\usepackage[colorlinks,allcolors=blue]{hyperref}

\jname{Environmental Data Science}
\articletype{RESEARCH ARTICLE}
\jyear{2026}

\begin{document}

\begin{Frontmatter}

\title[Characterizing AlphaEarth Embedding Geometry]{Characterizing AlphaEarth Embedding Geometry for \\
Agentic Environmental Reasoning}

\author[1]{Mashrekur Rahman}
\author[2]{Samuel J. Barrett}
\author[3]{Christina Last}

\authormark{Rahman \textit{et al}.}

\address[1]{\orgdiv{Dartmouth Libraries}, \orgname{Dartmouth College}, \orgaddress{\city{Hanover}, \postcode{03755}, \state{NH}, \country{USA}}. \email{mashrekur.rahman@dartmouth.edu}}

\address[2]{\orgname{LGND AI}, \orgaddress{\state{Canarias}, \country{Spain}}. \email{sam@lgnd.ai}}

\address[3]{\orgname{TipplyAI}, \orgaddress{\city{London}, \country{UK}}. \email{christina@tipply.ai}}

\keywords{satellite foundation models, embedding geometry, manifold characterization, agentic systems, retrieval-augmented generation, geospatial intelligence}

\abstract{Earth observation foundation models encode land surface information into dense embedding vectors, yet the geometric structure of these representations and its implications for downstream reasoning remain relatively less understood. An unexplored area is what operations can and should be performed on AlphaEarth embeddings, and how should geometric understanding inform the design of systems that reason over them? We address these gaps by characterizing the embedding space geometry of Google AlphaEarth's 64-dimensional embeddings across 12.1 million samples in the Continental United States (2017--2023) and develop an agentic system that leverages this geometric understanding for environmental reasoning. The manifold is non-Euclidean: principal component analysis demonstrates an effective dimensionality of 13.3 (participation ratio) from 64 raw dimensions, while maximum likelihood estimation yields a local intrinsic dimensionality of approximately 10. Tangent spaces rotate substantially across the manifold, with 84\% of sampled locations exhibiting tangent space angles exceeding 60\textdegree{} and local-global principal component alignment (mean $|\cos\theta| = 0.17$) approaching the random baseline of 0.125. Compositional vector arithmetic analogous to word embedding analogies yields poor precision across three experiments. Supervised linear probes trained at three spatial scales show that concept directions rotate substantially across the manifold, and compositional vector arithmetic using both PCA-derived and probe-derived directions yields poor precision in our experiments. Retrieval, by contrast, produces physically coherent results across most of the manifold, with local geometric features predicting retrieval coherence ($R^2 = 0.32$). Building on this characterization, we introduce an agentic geospatial intelligence system equipped with nine specialized tools (five retrieval-based and four geometry-aware) that decomposes natural language environmental queries into multi-step reasoning chains over a FAISS-indexed embedding database. A five-condition ablation study using 120 queries across three complexity tiers demonstrates that satellite embedding retrieval is the dominant contributor to response quality ($\mu = 3.79 \pm 0.90$ vs.\ $\mu = 3.03 \pm 0.77$ for parametric-only generation; scale 1--5), while the agentic architecture achieves its strongest performance on multi-step comparison queries ($\mu = 4.28 \pm 0.43$), a task class that deterministic pipelines cannot address. A cross-model benchmark comparing Claude Sonnet~4.5 and Opus~4.6 on queries drawn from the same set indicates that geometric tools reduce Sonnet's weighted score by 0.12 points but improve Opus's by 0.07, with Opus receiving substantially higher geometric grounding scores (3.38 vs.\ 2.64). This effect is most apparent on single-location queries, where Opus gains 0.58 points from geometric context while Sonnet loses 0.29. The finding suggests that the value of embedding space geometric characterization is not fixed but scales with the reasoning capability of the consuming model.}

\end{Frontmatter}

\section{Introduction} \label{sec:introduction}

Earth observation foundation models compress multispectral imagery into dense embedding vectors that serve as compact representations of the land surface \citep{bodnar2025foundation, xiao2025foundation, zhu2024foundations, fang2026earth}. Models such as Google AlphaEarth \citep{brown2025alphaearth, tollefson2025google}, Prithvi \citep{jakubik2024prithvi}, Scale-MAE \citep{reed2023scalemai}, and Clay \citep{clay2024foundation} have demonstrated strong performance on downstream tasks including land cover classification, change detection, and crop mapping \citep{mai2023opportunities, murakami2025alphaearth, liu2025beyond}. These models are typically evaluated on task accuracy, but the internal structure of their embedding spaces remains to be comprehensively understood. What geometric properties do these representations have, and what do those properties imply about which operations can and cannot be performed on the embeddings?

In natural language processing, word embeddings support vector arithmetic: adding and subtracting vectors produces semantically meaningful results because the embedding space is approximately linear and the relevant directions are globally consistent \citep{mikolov2013efficient, mikolov2013linguistic}. The manifold hypothesis posits that learned representations occupy low-dimensional manifolds embedded in higher-dimensional ambient spaces \citep{bengio2013representation}, but these manifolds need not be flat. \citet{ethayarajh2019contextual} showed that the geometry of contextual word embeddings varies substantially across layers and contexts. If AlphaEarth embeddings lie on a curved, heterogeneous manifold, compositional operations that assume linearity may not transfer directly, and the choice of reasoning strategy should be informed by the geometry itself.

Large Language Models (LLMs) augmented with specialized tools are being applied to geospatial tasks \citep{wang2024gpt, zhang2023geogpt, xu2025agentic, sun2026multiagent, chen2025empowering, feng2025earthagent}. The ReAct framework \citep{yao2023react} interleaves reasoning traces with tool calls, while Toolformer \citep{schick2023toolformer} demonstrated that language models can learn to invoke external APIs autonomously. Retrieval-augmented generation (RAG) grounds language model outputs in retrieved evidence rather than relying solely on parametric knowledge \citep{lewis2020retrieval}. These approaches have been applied to text and image retrieval, but satellite foundation model embeddings have not been used as retrieval backbones for environmental reasoning. A critical prerequisite for such an application is understanding the geometric conditions under which retrieval produces physically meaningful results.

\citet{rahman2026physically} established a dimension-level dictionary for AlphaEarth embeddings, showing that individual dimensions map onto specific environmental properties and that the full embedding space reconstructs most environmental variables. That work characterized what each dimension encodes individually. But knowing what each dimension means does not tell us how the dimensions relate to each other, whether the space is flat or curved, or which operations the geometry actually supports. In natural language processing (NLP), this distinction has been significant: the field moved from interpreting individual word embedding dimensions to studying the geometry of the full space, because geometry determines which downstream operations are valid \citep{ethayarajh2019contextual, mu2018allbutthetop}. The present work makes this transition for earth observation foundation model embeddings, asking how the 64 dimensions relate to each other geometrically and what that structure implies for systems that reason over the space.

We address four research questions:

\begin{enumerate}
    \item What is the geometric structure of the AlphaEarth embedding manifold, and how does local geometry relate to global structure?
    \item Which reasoning operations are geometrically viable in this embedding space? 
    \item Can an agentic system with geometry-aware tools extend satellite-grounded intelligence beyond single-step retrieval to multi-step reasoning?
    \item Does the utility of embedding space geometric metadata depend on the reasoning capability of the consuming language model?
\end{enumerate}

\section{Data and Prior Work} \label{sec:data}

We use the same dataset and embedding infrastructure described in \citet{rahman2026physically}. Briefly, Google AlphaEarth embeddings \citep{brown2025alphaearth, tollefson2025google} were extracted through the Google Earth Engine API \citep{gorelick2017google}  at individual grid points spaced 0.025\textdegree{} apart (approximately 2.75~km) across the Continental United States (CONUS), bounded by 125.0\textdegree W--66.5\textdegree W and 24.5\textdegree N--49.5\textdegree N. Data span seven annual embeddings (2017--2023), approximately 12.1 million 64-dimensional embedding vectors (denoted A00--A63). Each vector is co-located with 26 environmental variables from MODIS, PRISM, ERA5-Land, SRTM, SoilGrids, and NLCD.

\citet{rahman2026physically} established three results that the present work builds upon. First, combining linear (Spearman correlation), nonlinear (Random Forest), and attention-based (TabTransformer) methods, that work showed that individual embedding dimensions correlate with identifiable land surface properties, while the full embedding space reconstructs most environmental variables (14 of 26 variables exceed $R^2 > 0.90$; temperature and elevation approach $R^2 = 0.97$). Second, these relationships are spatially robust under $2^{\circ} \times 2^{\circ}$ block cross-validation (mean generalization gap $\Delta R^2 = 0.017$) \citep{roberts2017crossvalidation, ploton2020spatial} and temporally stable across all seven study years (mean inter-year correlation $\bar{r} = 0.963$). Third, a five-stage deterministic pipeline using FAISS-indexed \citep{johnson2019billion} retrieval over this embedding database translates natural language environmental queries into satellite-grounded assessments, achieving $\mu = 3.74 \pm 0.77$ (scale 1--5) on an LLM-as-Judge evaluation \citep{zheng2023judging} with 360 query--response cycles.

\citet{rahman2026physically} treated each embedding dimension independently, characterizing what individual dimensions encode. Dimension-level analysis is a necessary first step, but it does not tell us whether the space supports the compositional operations that make embeddings useful in practice. The present work takes the next step by characterizing the manifold geometry, testing whether vector arithmetic analogous to word embedding analogies \citep{mikolov2013efficient} transfers to this space, and building an agentic system whose design is informed by the geometric structure.

\section{Methods} \label{sec:methods}

Our analysis proceeds in three phases. We first characterize the geometric structure of the AlphaEarth embedding manifold (Section~\ref{sec:manifold}), asking whether the space is flat or curved, globally uniform or locally heterogeneous. This characterization motivates the second phase, which tests whether compositional operations such as vector arithmetic transfer to this space (Section~\ref{sec:arithmetic}), and whether retrieval produces physically coherent results across the manifold (Section~\ref{sec:coherence}). Together, these findings inform the design of an agentic system (Sections~\ref{sec:agent}--\ref{sec:evaluation}) that can work with, rather than against, the structure of the embedding space.

\begin{figure}[t]%
\FIG{\includegraphics[width=0.9\textwidth]{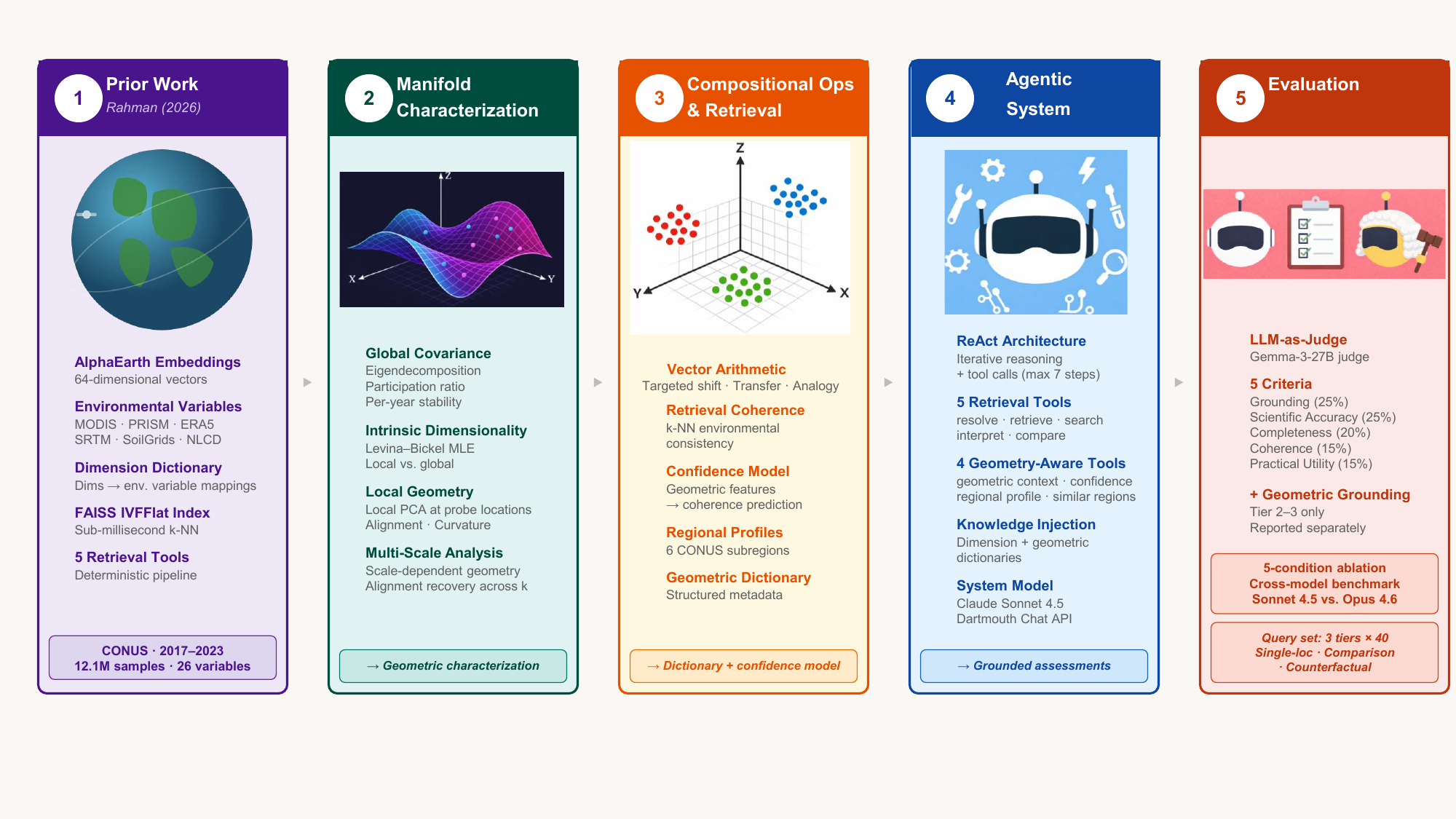}}
{\caption{Overview of the analysis and system architecture. (1)~Prior work \citep{rahman2026physically} established a dimension dictionary mapping AlphaEarth's 64 embedding dimensions to environmental variables and built a FAISS-indexed retrieval pipeline over 12.1 million CONUS samples. (2)~Phase~1 characterizes the manifold geometry through global covariance analysis, intrinsic dimensionality estimation, local PCA, and multi-scale analysis. (3)~Phase~2 tests compositional vector arithmetic and supervised linear probes across spatial scales, measures retrieval coherence across the manifold, and builds a confidence model and regional geometric profiles. (4)~The agentic system combines five retrieval tools from \citet{rahman2026physically} with four new geometry-aware tools under a ReAct-style planning architecture. (5)~Evaluation uses an LLM-as-Judge framework with five weighted criteria across 120 queries in three complexity tiers, with a five-condition ablation and a cross-model benchmark comparing Claude Sonnet~4.5 and Opus~4.6}
\label{fig:architecture}}
\end{figure}

\subsection{Phase 1: Manifold Characterization} \label{sec:manifold}

\citet{rahman2026physically} characterized what individual embedding dimensions encode. Here we ask a different question: how do the 64 dimensions relate to \emph{each other}, and what does the embedding space look like as a geometric object?

\paragraph{Global covariance structure.}
We begin at the largest scale. If the embedding space were organized along a small number of independent axes, most variance would concentrate in a few principal components; if all 64 dimensions contributed independently, the space would be uniformly 64-dimensional. To determine which regime applies, we computed the $64 \times 64$ covariance and Spearman rank correlation matrices from a balanced subsample of one million vectors (approximately 143,000 per year across 2017--2023) and performed eigendecomposition. We quantified effective dimensionality using the participation ratio \citep{gao2017theory}:
\begin{equation}
    \text{PR} = \frac{\left(\sum_i \lambda_i\right)^2}{\sum_i \lambda_i^2}
    \label{eq:pr}
\end{equation}
where $\lambda_i$ are the eigenvalues. The participation ratio equals 1 when a single component dominates and equals $d$ when all $d$ components contribute equally, providing a continuous measure of how many dimensions carry meaningful variance. We cross-referenced the eigenvector loadings with the dimension--variable assignments from \citet{rahman2026physically} to determine whether the principal axes of variation correspond to recognizable environmental gradients (e.g., a moisture axis, a temperature axis).

To test whether this geometric structure is stable or changes from year to year, we repeated the eigendecomposition independently for each of the seven years and computed pairwise principal subspace angles. We also applied hierarchical clustering (Ward's method) on the absolute correlation matrix to identify groups of co-varying dimensions.

\paragraph{Intrinsic dimensionality.}
The participation ratio characterizes the global shape of the eigenspectrum, but says nothing about local complexity. A manifold can have low global dimensionality yet be locally folded or twisted in ways that make the neighborhood of each point more complex than the global average suggests, or conversely, locally simpler if each neighborhood occupies a low-dimensional subspace. To capture this, we estimated local intrinsic dimensionality at each point using the maximum likelihood estimator of \citet{levina2004maximum}:
\begin{equation}
    \hat{d}_k(x) = \left[\frac{1}{k-1} \sum_{j=1}^{k-1} \log \frac{r_k(x)}{r_j(x)}\right]^{-1}
    \label{eq:mle_id}
\end{equation}
where $r_j(x)$ is the Euclidean distance from point $x$ to its $j$-th nearest neighbor. Intuitively, this estimator measures how quickly the volume of the neighborhood grows as the radius increases: faster growth implies more local dimensions. We applied it to 200,000 locations (balanced across years) at neighborhood sizes $k \in \{5, 10, 20, 30, 50, 75, 100\}$ and mapped the spatial distribution of local intrinsic dimensionality across CONUS. Per-year estimates (100,000 points each) assessed temporal stability. To visualize the manifold, we projected the 200,000 points into three dimensions using PCA and colored them by local intrinsic dimensionality and by elevation, revealing how geometric complexity relates to physical geography. We additionally stratified local intrinsic dimensionality by elevation bands to test whether topographic complexity corresponds to geometric complexity in the embedding space (results in Figure~\ref{fig:intrinsic_dim}).

\paragraph{Local geometry and tangent spaces.}
The intrinsic dimensionality analysis tells us \emph{how many} local directions matter, but not \emph{which} directions they are. A critical question for downstream applications is whether the global principal components, which \citet{rahman2026physically} identified as moisture--vegetation and temperature axes, also describe local variation. If they do, a single global dictionary suffices; if they do not, the embedding space is heterogeneous and requires spatially varying interpretation.

We addressed this by performing local PCA at 10,000 probe locations stratified by elevation, using $k = 100$ nearest neighbors per probe. For each probe we computed three quantities. First, the local participation ratio from the neighborhood eigenspectrum, measuring how many directions carry local variance. Second, alignment between the local and global first principal components, measured as $|\cos\theta|$ between the local PC1 and the global PC1. For reference, the expected alignment under random rotation in 64 dimensions is $\mathbb{E}[|\cos\theta|] = \sqrt{2/(\pi \cdot 64)} \approx 0.125$; values near this baseline indicate that local and global principal directions are unrelated. Third, tangent space instability, measured as the principal angle between the tangent spaces of adjacent probes, capturing how rapidly the local geometry changes across the manifold.

We additionally recorded the dominant environmental category (from the dimension dictionary of \citet{rahman2026physically}) at each probe to map which physical processes, whether temperature, vegetation, hydrology, soil, or terrain, dominate local variation at each location. To examine whether flatter regions of the manifold also tend to align better with global structure, we analyzed the joint distribution of alignment and tangent angle across all probes.

\paragraph{Multi-scale analysis.}
The local PCA results characterize geometry at a single neighborhood size ($k = 100$, corresponding to roughly 25--50~km depending on local point density). But the relationship between local and global geometry may depend on scale: local axes might align poorly with global ones at small scales but gradually converge at larger scales as the neighborhood averages over more diverse environments. To test this, we repeated the full local PCA analysis at the same 10,000 probes for neighborhood sizes $k \in \{20, 100, 500, 2000\}$ and tracked five quantities as functions of scale: local--global PC1 alignment, local--global PC2 alignment, tangent space angle, local participation ratio, and the fraction of variance explained by the first local principal component. We also recorded how the dominant environmental category shifts across scales, testing whether, for example, terrain-dominated local variation at small scales gives way to climate-dominated variation at continental scales. The scale at which alignment recovers, if it does, defines the minimum spatial extent over which the global dimension dictionary from \citet{rahman2026physically} can be applied reliably.

\subsection{Phase 2: Compositional Operations} \label{sec:arithmetic}

The manifold characterization determines whether the embedding space has the geometric properties needed for compositional reasoning. In natural language processing, word embeddings famously support vector arithmetic: $\vec{\text{king}} - \vec{\text{man}} + \vec{\text{woman}} \approx \vec{\text{queen}}$ \citep{mikolov2013efficient, mikolov2013linguistic}. This works because word embedding spaces are approximately linear and the relevant semantic axes are globally consistent. If the AlphaEarth manifold is similarly structured, one could construct synthetic embeddings by shifting along interpretable directions, for instance making a location ``wetter'' or ``hotter'' through vector addition. We tested this through three experiments of increasing complexity, using two methods for identifying concept directions. The first method selects the single local principal component most correlated with the target property, isolating whether any individual axis of local variation aligns with the target. The second method trains supervised linear probes (Ridge regression) that combine all 64 dimensions to find an optimal concept direction. Both methods are applied to each experiment, allowing us to separate whether arithmetic failures stem from crude direction estimates or from the geometry of the space itself.

\paragraph{Targeted shift.}
The simplest compositional operation is a single-property shift: given a location's embedding, move it along the direction associated with a target property. For each of six target properties (precipitation, temperature, NDVI, elevation, soil organic carbon, evapotranspiration), we performed local PCA at each of 500 source locations using $k = 100$ neighbors. We then computed the Pearson correlation between each of the top 10 local principal components and the target environmental variable across the neighborhood, and selected the single PC with the highest absolute correlation as the shift direction.

This single-PC selection is a deliberate simplification: a linear probe (regression or lasso across multiple PCs) could in principle identify a better concept direction by combining several axes, but the single-PC approach isolates whether any individual axis of local variation aligns with the target property, which is the more conservative test of compositional structure. We shifted each source embedding by $n\sigma$ along the selected direction,

where $\sigma$ is the standard deviation of the local neighbor embeddings projected onto that direction and $n \in \{0.5, 1.0, 1.5, 2.0\}$, then retrieved the nearest real embedding to the shifted point via FAISS. We evaluated two quantities: the change in the target property (did the shift achieve the intended effect?) and collateral changes in non-target properties (did other properties remain stable?). We compared four shift methods: global direction (the global PC with highest target correlation, computed over 50,000 samples), local direction (the local PC described above), random direction (a unit vector drawn uniformly at random in 64 dimensions), and a geographic baseline (the nearest real location with the desired property change).

\paragraph{Property transfer.}
A more demanding operation is transplanting a specific property from one location to another. For example, given a forested mountain location A and a wet lowland location B, we asked whether we could produce an embedding that preserves A's terrain and vegetation characteristics but adopts B's precipitation regime. To do this, we performed local PCA at A, identified the local principal components most correlated with the target property (precipitation), replaced those components with the corresponding values from B's embedding, and left the remaining components unchanged. This succeeds only if the target property occupies a separable subspace of the local tangent space. We evaluated target error (how closely the result matches B's target property) and non-target deviation (how much A's other properties changed) in units of the property standard deviation.

\paragraph{Analogy.}
In NLP, $\vec{\text{king}} - \vec{\text{man}} + \vec{\text{woman}} \approx \vec{\text{queen}}$ works because the gender direction is globally consistent. We tested the geospatial analogue: given three locations A, B, and C where A and B differ primarily in one environmental property (e.g., A is dry and B is wet, but both share similar terrain), we computed $\vec{A} - \vec{B} + \vec{C}$ and asked whether the resulting vector $\vec{D}$ lands near a real location that resembles C but with B's value of the target property.  We tested two variants: a na\"ive approach that applies the offset $\vec{B} - \vec{A}$ directly in ambient space, and a local approach that projects the offset onto C's tangent space before applying it. In both cases, we retrieved the nearest real embedding to $\vec{D}$ via FAISS and compared its environmental profile against the expected outcome.

\paragraph{Linear probes for concept directions.}
An alternative to PCA is to train supervised linear probes that combine information across all embedding dimensions to find an optimal concept direction for each property. 
To test whether supervised directions rescue compositional arithmetic, we trained Ridge regression models ($\alpha = 1.0$) predicting each target property from the 64-dimensional embedding vector at three spatial scales: (1)~global, using all CONUS samples; (2)~regional, within each of five CONUS subregions; and (3)~local, using the $k = 100$ nearest neighbors at each of the same 500 source locations. For each probe we extracted the normalized coefficient vector as the concept direction and measured two quantities: predictive accuracy ($R^2$) and direction stability (cosine similarity between local, regional, and global directions for the same property). We then repeated the targeted shift experiment using probe-derived directions at all three scales, with PCA-derived local directions and random directions as baselines, evaluating the same target change and non-target preservation metrics.

\subsection{Retrieval Coherence and Geometric Dictionary} \label{sec:coherence}

The arithmetic experiments test whether we can \emph{construct} new valid embeddings through algebraic manipulation. But the primary operation in retrieval-augmented generation is not construction; it is \emph{retrieval}: given a query embedding, find the nearest neighbors in the database and use their associated metadata. Even if arithmetic fails, retrieval may still work well, provided that nearby embeddings in the database encode similar physical environments. The question is whether this holds uniformly across the manifold or whether some regions produce more coherent retrievals than others.

\paragraph{Retrieval coherence.}
We measured retrieval coherence at 10,000 locations by computing the coefficient of variation of each environmental variable among the $k = 10$ nearest neighbors in embedding space, then averaging across variables to obtain a single coherence score per location. A low score indicates that the retrieved neighbors share similar environmental properties; a high score indicates that embedding proximity does not reliably correspond to physical similarity, and downstream responses grounded in such retrievals may be unreliable.

\paragraph{Confidence model.}
If retrieval coherence varies spatially, can it be predicted from the geometric features we have already computed? We fit a linear model predicting coherence from five geometric features: local intrinsic dimensionality, local participation ratio, mean embedding distance to neighbors, tangent space angle, and local--global PC1 alignment. A predictive relationship would mean the agent system can estimate retrieval reliability at any location \emph{before} examining the ground-truth environmental data, providing a built-in confidence score.

\paragraph{Regional profiles and geometric dictionary.}
To make the spatial heterogeneity of the embedding space actionable for the agent, we partitioned CONUS into six subregions (Great Plains, Southeast, Northeast, Mountain West, Southwest, Pacific Northwest) and computed regional profiles documenting three quantities: mean retrieval coherence, the embedding dimensions with highest local importance (from local PCA loadings), and mean local intrinsic dimensionality. These regional profiles, combined with the per-dimension spatial importance maps and confidence model coefficients, constitute the \emph{enhanced geometric dictionary}: a structured metadata layer that tells the agent system which dimensions matter most in each region and how much to trust its retrievals at any given location.

\subsection{Phase 3: Agentic System} \label{sec:agent}

The geometric characterization and retrieval coherence analysis establish what operations are viable in the embedding space and where those operations are reliable. We now describe the agentic system designed to use these findings for environmental reasoning.

\paragraph{Architecture.}
The system follows a ReAct-style architecture \citep{yao2023react} in which a language model alternates between reasoning traces and tool calls. Given a natural language query, the agent first generates a reasoning trace that decomposes the query into subgoals, then selects and invokes tools to address each subgoal, observes the results, and iterates until it can synthesize a final response. The maximum number of planning--execution iterations is capped at seven per query to prevent runaway loops.

\paragraph{Tool set.}
The agent has access to nine callable tools organized into two groups. Five tools are adapted from the deterministic pipeline of \citet{rahman2026physically}: (1)~\texttt{resolve\_location}, which geocodes natural language place names to coordinates within CONUS; (2)~\texttt{retrieve\_embedding}, which fetches the 64-dimensional embedding vector and 26 co-located environmental variables for a given coordinate and year from the FAISS-indexed database of 12.1 million vectors; (3)~\texttt{search\_similar}, which performs $k$-nearest-neighbor search in embedding space to find physically similar locations; (4)~\texttt{interpret\_dimensions}, which looks up the physical meaning of each dimension using the dimension dictionary from \citet{rahman2026physically}; and (5)~\texttt{compare\_locations}, which retrieves and contrasts the embedding profiles of two locations.

Four new tools expose the geometric characterization from Phases 1--2: (6)~\texttt{get\_geometric\_context}, which returns local intrinsic dimensionality, the dominant environmental category, the predicted retrieval confidence, and the regional profile for a location; (7)~\texttt{assess\_retrieval\_confidence}, which applies the calibrated confidence model to predict retrieval coherence at a location before retrieval occurs; (8)~\texttt{get\_regional\_profile}, which returns the geometric profile (dominant dimensions, mean coherence, local intrinsic dimensionality) for a named CONUS subregion; and (9)~\texttt{identify\_similar\_regions}, which ranks subregions by geometric similarity to a query location.

\paragraph{System model.}
All tool-call planning and response synthesis is performed by the system model (Claude Sonnet~4.5) \citep{anthropic2025claude}, accessed through an OpenAI-compatible institutional API endpoint \citep{dartmouth_chat}. The system model receives the tool manifest, query, and accumulated observations at each reasoning step, and generates the next tool call or final response.

\subsection{Evaluation} \label{sec:evaluation}

\paragraph{Query set.}
We designed 120 evaluation queries organized into three tiers of increasing complexity, with 40 queries per tier. Tier~1 queries request single-location environmental assessments (e.g., ``Describe the land surface near Wichita, Kansas''), directly overlapping with the \citet{rahman2026physically} evaluation set to enable benchmarking. Tier~2 queries require multi-step comparison across two or more locations (e.g., ``Compare flood risk between Portland, Oregon and Phoenix, Arizona''), requiring the agent to resolve multiple locations, retrieve embeddings for each, and synthesize a comparative response. Tier~3 queries require context-aware or counterfactual reasoning that benefits from geometric awareness (e.g., ``Assess drought vulnerability across the Great Plains, considering how retrieval confidence varies with local terrain complexity''). Queries are distributed across all six CONUS subregions and span multiple intent categories (location profiling, risk assessment, vegetation analysis, soil properties, climate, hydrology, and urban characterization).

\paragraph{Ablation conditions.}
We evaluate five conditions: (1)~\emph{Full system}: all nine tools available; (2)~\emph{No geometric}: only the five \citet{rahman2026physically} tools, removing all geometry-aware tools; (3)~\emph{No confidence}: geometric context available but retrieval confidence scoring disabled; (4)~\emph{Deterministic}: the \citet{rahman2026physically} deterministic pipeline (no LLM-based planning), which can only process Tier~1 queries; (5)~\emph{LLM only}: the language model generates responses without any tool access or retrieval, relying solely on parametric knowledge. All agentic conditions (1--3, 5) process all 120 queries; the deterministic condition processes only the 40 Tier~1 queries.

\paragraph{LLM-as-Judge scoring.}
Each response is scored by a separate judge model (Gemma-3-27B) \citep{gemma2025technical} on five criteria using a 1--5 scale: grounding (does the response cite retrieved data?), scientific accuracy (are interpretations consistent with validated relationships?), completeness (does it fully address the query?), coherence (is it well-structured and internally consistent?), and practical utility (is the information actionable?). The weighted score combines these as $0.25 \cdot G + 0.25 \cdot A + 0.20 \cdot C + 0.15 \cdot H + 0.15 \cdot U$, matching the weighting used in \citet{rahman2026physically} for comparability. An additional criterion, geometric grounding (GG), is scored for Tier~2 and Tier~3 queries to measure whether responses reference manifold properties such as confidence, dimensionality, or regional context. GG is reported separately and not included in the weighted score to preserve comparability with \citet{rahman2026physically}. The system model and judge model are always different to prevent self-evaluation bias \citep{zheng2023judging}.

\paragraph{Cross-model benchmark.}
To test whether the utility of geometric metadata depends on model capability, we re-ran the Full and No-geometric conditions on a subset of 36 queries (12 per tier) using Claude Opus~4.6 \citep{anthropic2025claude} as the system model, with the same Gemma-3-27B judge. All other parameters (tools, query set, scoring protocol) are held constant, isolating the effect of system model capability on geometric context utilization.

\section{Results} \label{sec:results}

We present results organized around the four research questions posed in Section~\ref{sec:introduction}. We first characterize the manifold geometry (RQ1), then report the outcome of compositional arithmetic experiments (RQ2), followed by the agentic system evaluation (RQ3) and the cross-model benchmark (RQ4).

\subsection{RQ1: What Is the Geometric Structure of the Embedding Manifold?}

\begin{figure}[t]%
\FIG{\includegraphics[width=0.9\textwidth]{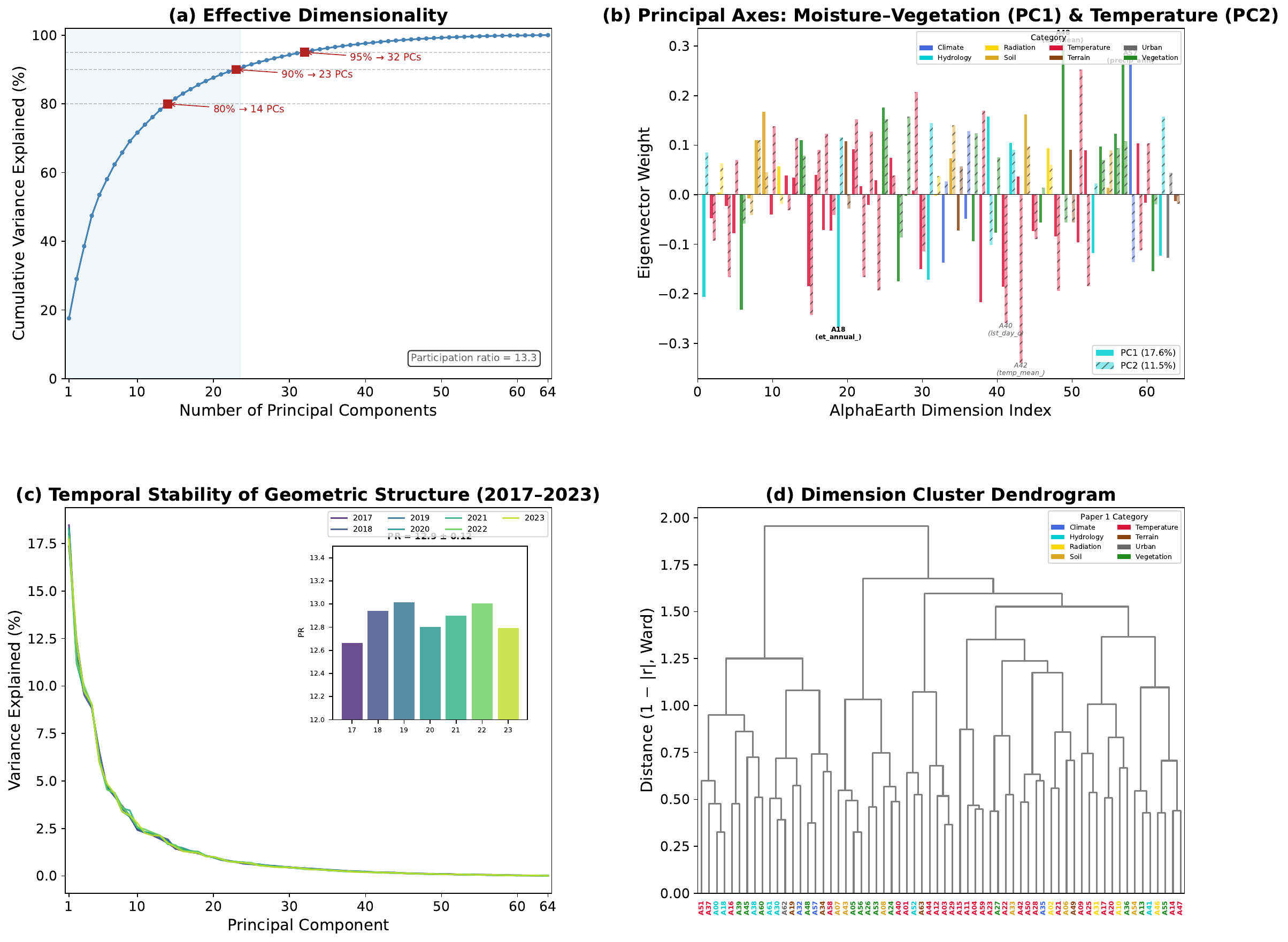}}
{\caption{Geometric structure of the AlphaEarth embedding space. (a) Cumulative variance explained by principal components, with the participation ratio of 13.3 annotating the effective dimensionality. Markers indicate the number of components needed for 80\%, 90\%, and 95\% of total variance. (b) Eigenvector weights for the top two principal components (PC1: moisture--vegetation, PC2: temperature), colored by \citet{rahman2026physically}'s environmental category assignments. (c) Per-year eigenvalue spectra (2017--2023) overlaid, showing temporal stability of the geometric structure. (d) Dendrogram from hierarchical clustering (Ward's method) of the $64 \times 64$ absolute correlation matrix, with leaf colors indicating \citet{rahman2026physically} environmental categories}
\label{fig:global_geometry}}
\end{figure}

\paragraph{Global structure.}
Eigendecomposition of the $64 \times 64$ covariance matrix reveals that the embedding space is organized along a moderate number of principal axes. The participation ratio is 13.3, indicating that the 64 raw dimensions effectively span approximately 13 independent directions of variation. The first principal component (17.6\% of variance) loads most heavily on moisture and vegetation dimensions: A48 (EVI, $\rho = +0.73$), A57 (precipitation, $\rho = +0.78$), A56 (LAI), and A18 (ET, $\rho = -0.67$). The second component (11.5\%) captures a temperature gradient: A40 (LST, $\rho = +0.78$), A42 (temperature, $\rho = +0.68$), A50 (temperature, $\rho = -0.71$). The principal axes of the embedding space thus correspond to the dominant environmental gradients across CONUS (Figure~\ref{fig:global_geometry}a,b).

This geometric structure is temporally stable. Per-year participation ratios are $12.9 \pm 0.12$ across the seven years, and per-year eigenvalue spectra are visually indistinguishable (Figure~\ref{fig:global_geometry}c). Hierarchical clustering identifies a silhouette peak at $k = 22$ clusters with a score of 0.147, indicating that dimensions form a continuum of co-variation rather than discrete groups. Of the 2,016 possible dimension pairs, 47 exceed $|r| > 0.5$, suggesting that the space is moderately disentangled: most dimensions are not redundant, but some share substantial variance (Figure~\ref{fig:global_geometry}d).

\begin{figure}[t]%
\FIG{\includegraphics[width=0.9\textwidth]{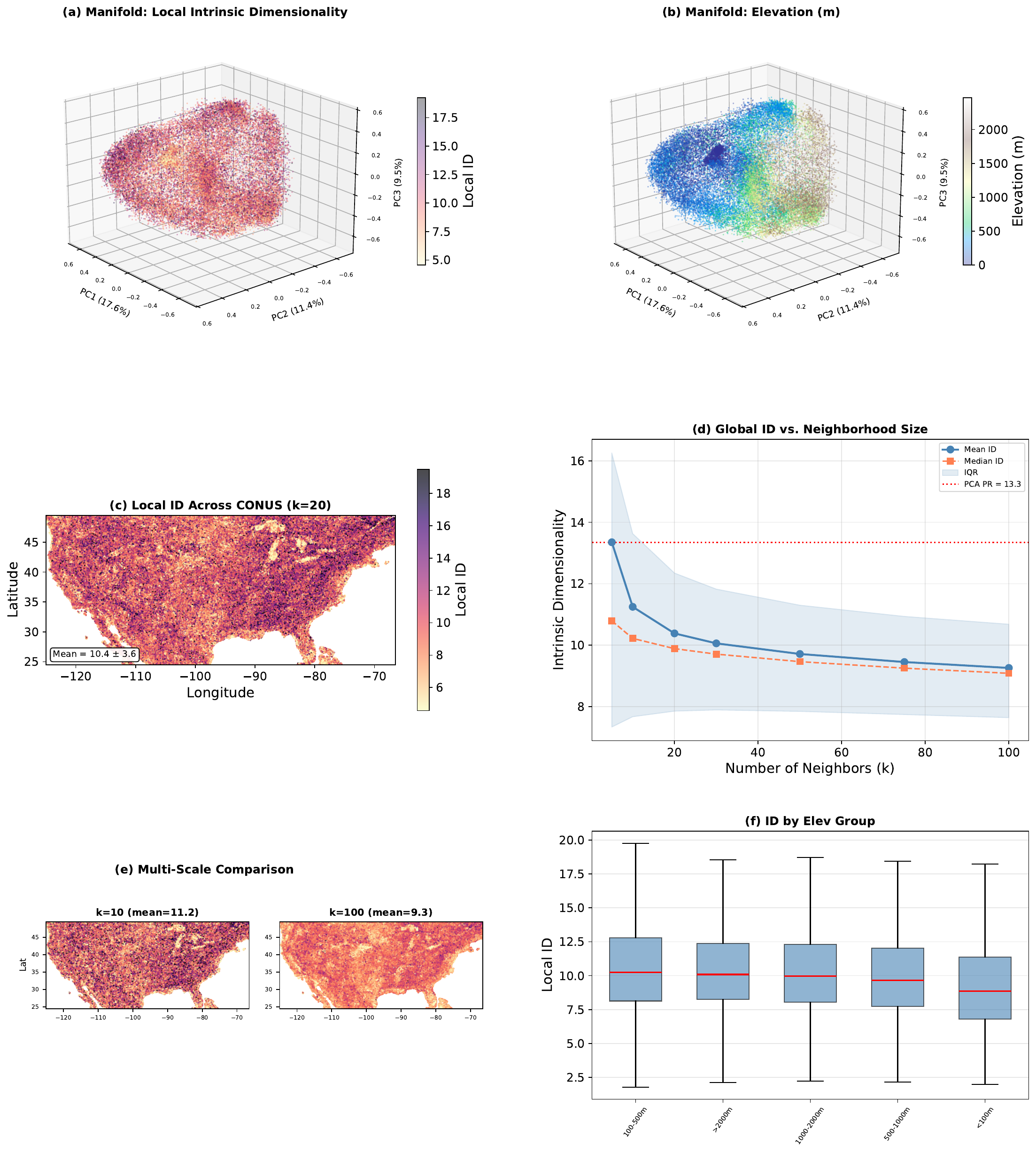}}
{\caption{Intrinsic dimensionality of the AlphaEarth embedding manifold. (a) Three-dimensional PCA projection of 200,000 embedding vectors colored by local intrinsic dimensionality (Levina--Bickel MLE, $k = 20$). (b) Same projection colored by elevation, showing that high-dimensionality regions correspond to topographically complex terrain. (c) Spatial distribution of local intrinsic dimensionality across CONUS (mean $= 10.4 \pm 3.6$). (d) Global intrinsic dimensionality as a function of neighborhood size $k$, with the PCA participation ratio (13.3) shown for reference; the gap between MLE and PR indicates manifold curvature. (e) Multi-scale comparison of local ID maps at $k = 10$ and $k = 100$. (f) Local intrinsic dimensionality stratified by elevation group}
\label{fig:intrinsic_dim}}
\end{figure}

\paragraph{Intrinsic dimensionality.}
The Levina--Bickel MLE estimator yields a mean local intrinsic dimensionality of $10.4 \pm 3.6$ at $k = 20$, dropping to $9.3$ at $k = 100$ (Figure~\ref{fig:intrinsic_dim}c,d). The ratio of MLE intrinsic dimensionality to participation ratio ($10.4 / 13.3 = 0.78$) is less than one, indicating that the manifold is genuinely curved: a flat manifold would yield a ratio near 1.0. Intrinsic dimensionality varies spatially across CONUS (Figure~\ref{fig:intrinsic_dim}a,b). Mountain fronts and biome transition zones exhibit the highest local dimensionality, while the Great Plains shows the lowest, consistent with the expectation that environmentally homogeneous regions occupy simpler submanifolds. When stratified by elevation (Figure~\ref{fig:intrinsic_dim}f), locations above 2,000~m show modestly higher intrinsic dimensionality than lowland areas, possibly from the additional terrain-driven complexity at high elevations. Per-year estimates are stable, indicating that the dimensionality structure is a persistent property of the embedding space.

This range is consistent with \citet{rao2025intrinsic}, who reported intrinsic dimensions between 2 and 10 for geographic implicit neural representations with ambient dimensions of 256 to 512. AlphaEarth's local ID of approximately 10 in a 64-dimensional ambient space suggests a comparatively higher ratio of intrinsic to ambient dimensionality.

\begin{figure}[t]%
\FIG{\includegraphics[width=0.9\textwidth]{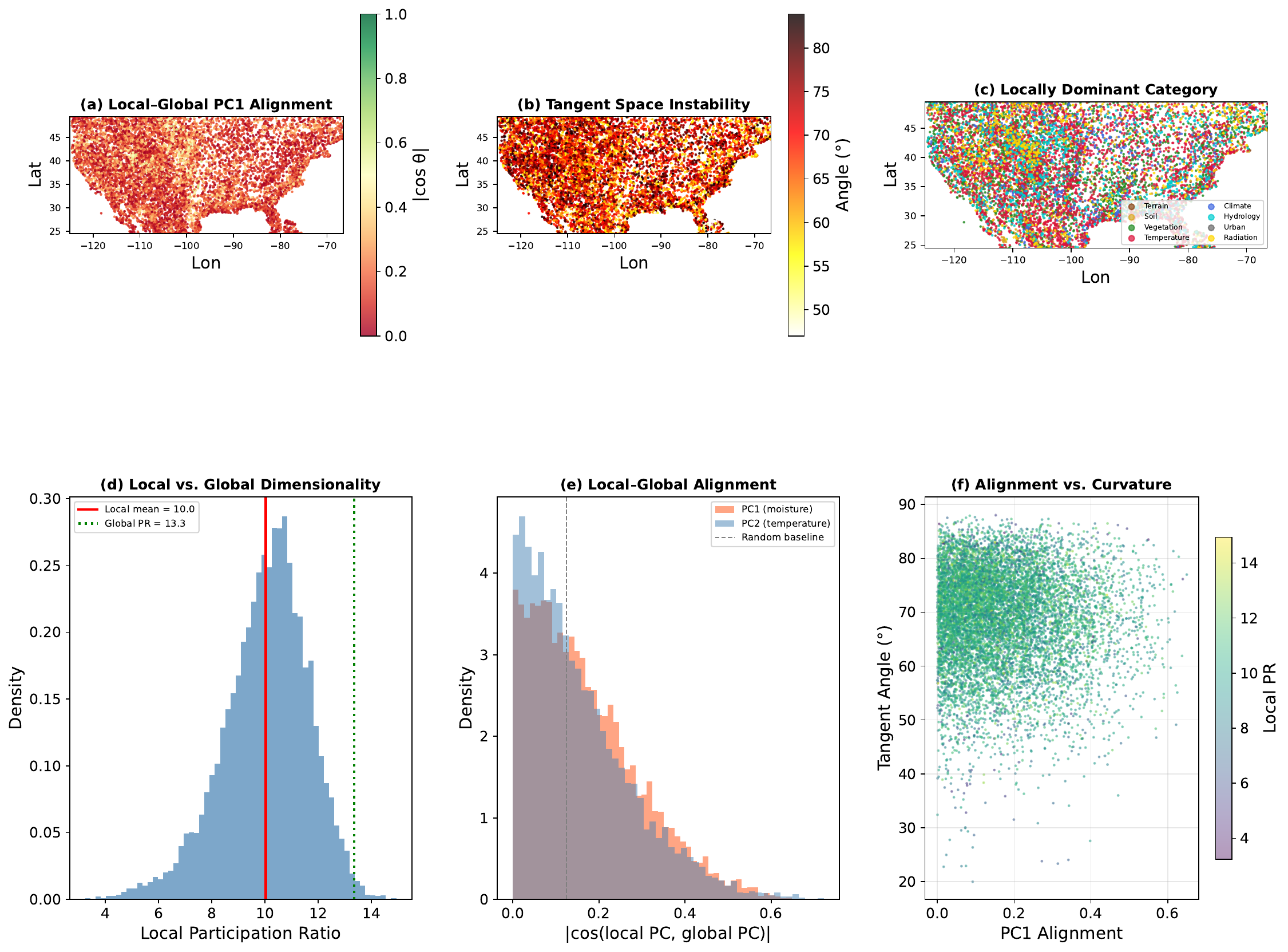}}
{\caption{Local geometry and tangent space analysis at 10,000 probe locations ($k = 100$ neighbors). (a) Alignment between local and global PC1 ($|\cos\theta|$) across CONUS; warm colors indicate high alignment. (b) Tangent space instability measured as the angle between adjacent tangent spaces; 84\% of locations exceed 60\textdegree. (c) Locally dominant environmental category at each probe, showing that temperature dominates 36\% of locations and vegetation 23\%. (d) Distribution of local participation ratios (mean $= 10.0$) compared to the global PR of 13.3. (e) Distribution of local--global PC1 alignment for PC1 (moisture) and PC2 (temperature), with the random baseline at 0.125. (f) Joint distribution of alignment and curvature, showing that locations with high tangent angles tend to have low alignment with global structure}
\label{fig:local_geometry}}
\end{figure}

\paragraph{Local geometry.}
Local PCA at 10,000 probe locations reveals that the manifold is heavily curved and locally heterogeneous (Figure~\ref{fig:local_geometry}). The mean alignment between local and global first principal components is $|\cos\theta| = 0.169$, barely above the random baseline of 0.125 (Figure~\ref{fig:local_geometry}e). This means that the moisture--vegetation axis identified globally by PC1 does not describe the primary direction of local variation at most locations (Figure~\ref{fig:local_geometry}a). Of the 10,000 probes, 84\% exhibit tangent space angles exceeding 60\textdegree{} between adjacent probes (Figure~\ref{fig:local_geometry}b), suggesting that the tangent space rotates substantially across the manifold. Spatially, the few locations with higher alignment cluster in the Great Plains and parts of the Southeast (Figure~\ref{fig:local_geometry}a), where the landscape is 
relatively flat and the global moisture-vegetation axis describes local variation reasonably well. The highest tangent space instability occurs in the mountainous West (Figure~\ref{fig:local_geometry}b), where steep environmental gradients produce rapidly changing local geometry.

Local participation ratios average 10.0, lower than the global value of 13.3 (Figure~\ref{fig:local_geometry}d), indicating that individual neighborhoods are simpler than the space as a whole. The dominant environmental category varies geographically (Figure~\ref{fig:local_geometry}c): temperature-related dimensions dominate 36\% of probe locations, vegetation 23\%, hydrology 12\%, and soil 10\%. The joint distribution of alignment and tangent angle (Figure~\ref{fig:local_geometry}f) shows that locations with high curvature also tend to have low alignment with global structure, as expected for a manifold whose principal directions rotate across its extent. We note that this heterogeneity may partly reflect the fact that Earth surface processes themselves operate at different scales: terrain and soil dominate local variation while temperature and moisture gradients dominate at continental scales. The embedding space inherits this multi-scale structure from its training data. The practical consequence is the same regardless of origin: a single global dictionary cannot describe local variation, and any system reasoning over these embeddings must account for spatially varying geometry.

\begin{figure}[t]%
\FIG{\includegraphics[width=0.9\textwidth]{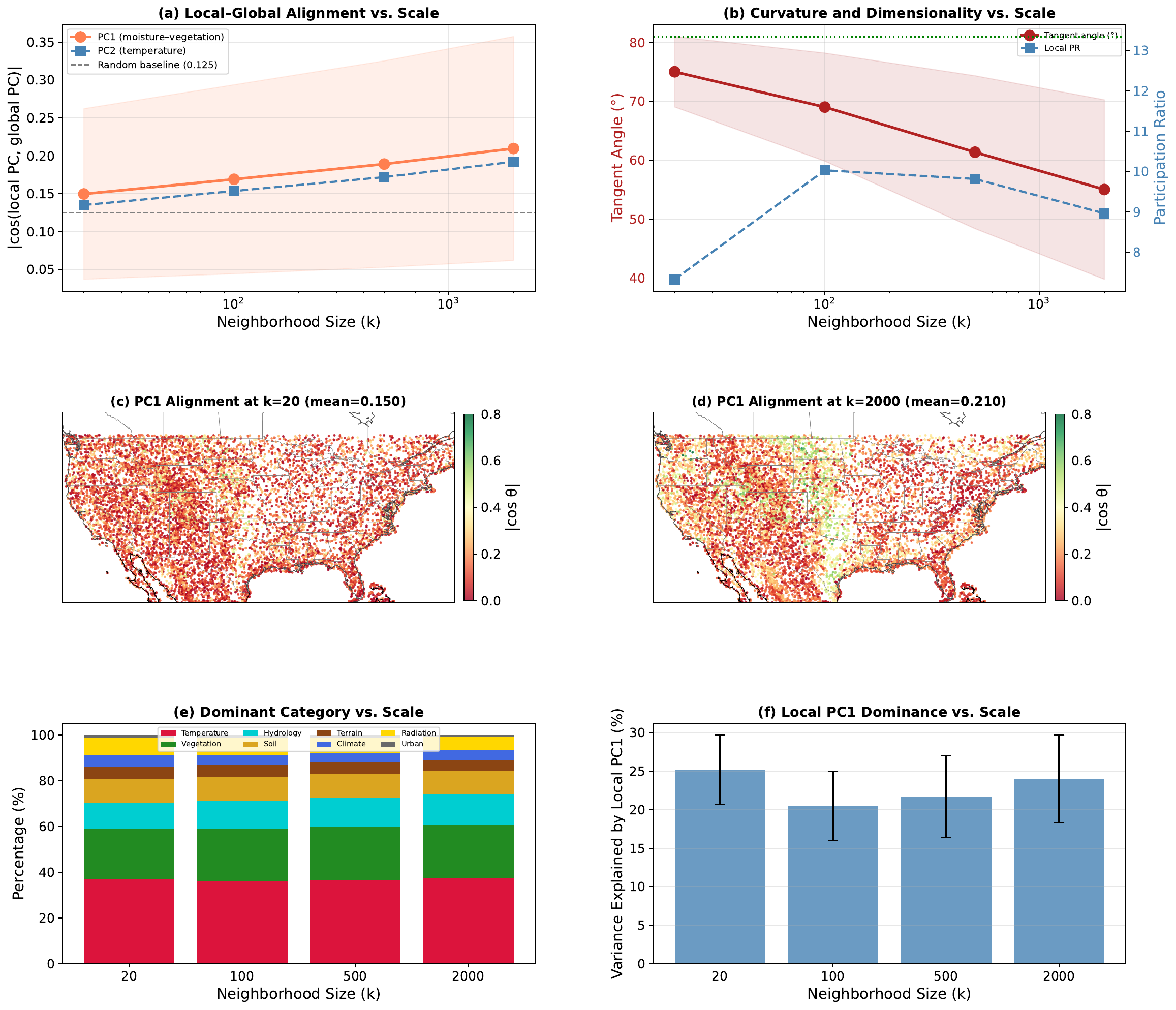}}
{\caption{Multi-scale geometric analysis at neighborhood sizes $k \in \{20, 100, 500, 2000\}$. (a) Local--global alignment for PC1 (moisture--vegetation) and PC2 (temperature) as a function of $k$; alignment increases slowly but remains below 0.3 even at $k = 2{,}000$. (b) Tangent angle and local participation ratio as functions of $k$. (c) Spatial map of PC1 alignment at $k = 20$ (mean $= 0.150$). (d) Spatial map of PC1 alignment at $k = 2{,}000$ (mean $= 0.210$). (e) Dominant environmental category as a function of scale, showing the transition from terrain/soil dominance at small $k$ to temperature/climate dominance at large $k$. (f) Variance explained by the first local principal component as a function of $k$, decreasing as larger neighborhoods average over more diverse environments}
\label{fig:multiscale}}
\end{figure}

\paragraph{Multi-scale behavior.}
Repeating the local PCA at neighborhood sizes from $k = 20$ to $k = 2{,}000$ reveals that local--global alignment increases only gradually with scale: 0.150 at $k = 20$ to 0.210 at $k = 2{,}000$ (Figure~\ref{fig:multiscale}a). Spatial maps of PC1 alignment (Figure~\ref{fig:multiscale}c,d) show that alignment is uniformly low at small scales and improves modestly across the interior at large scales, though no region approaches the value of 1.0 expected for a globally consistent space. Even at the largest neighborhoods, alignment does not exceed 0.3. The dominant environmental category shifts across scales (Figure~\ref{fig:multiscale}e), with terrain and soil giving way to temperature and climate at larger neighborhoods, reflecting the transition from local topographic variation to continental-scale climatic gradients. The fraction of variance explained by the first local principal component also decreases with increasing $k$ (Figure~\ref{fig:multiscale}f), indicating that larger neighborhoods average over more diverse environments and no single axis dominates. Together, these results indicate that the global dimension dictionary from \citet{rahman2026physically} is a useful but locally approximate description of the embedding space.

\subsection{RQ2: Can Compositional Operations Be Performed in This Space?}

\paragraph{Arithmetic experiments.}
All three compositional experiments using PCA-derived directions yield poor results. In the targeted shift experiment, global directions perform barely above the random baseline in producing the intended property change, while local directions achieve somewhat better precision but introduce substantial collateral changes in non-target properties. Property transfer and analogy experiments similarly produce embeddings that do not match the expected property profiles.

\paragraph{Linear probes and direction stability.}
If the failure of vector arithmetic is merely a consequence of using crude PCA-derived directions, then supervised concept directions should rescue it. We tested this by training Ridge regression probes that 
predict each target property from the full 64-dimensional embedding vector. The probes fit well: globally, $R^2$ ranges from $0.84$ (precipitation) to $0.98$ (elevation). At the local scale ($k = 100$ neighbors), precipitation probes achieve a median $R^2 = 0.93$, exceeding the global fit. At the 0.025\textdegree{} grid spacing used here, $k = 100$ neighbors span approximately 25 to 50~km depending on local point density. Precipitation varies relatively little over such distances in most of CONUS, which contributes to the high local $R^2$: the prediction task is easier when the target variable has a narrow range within the neighborhood. The relevant finding is not that local probes fit well, but that the directions they identify rotate across the manifold despite this high fit. The embeddings contain the relevant information, and the probes extract it to an extent.

\begin{figure}[t]%
\FIG{\includegraphics[width=0.9\textwidth]{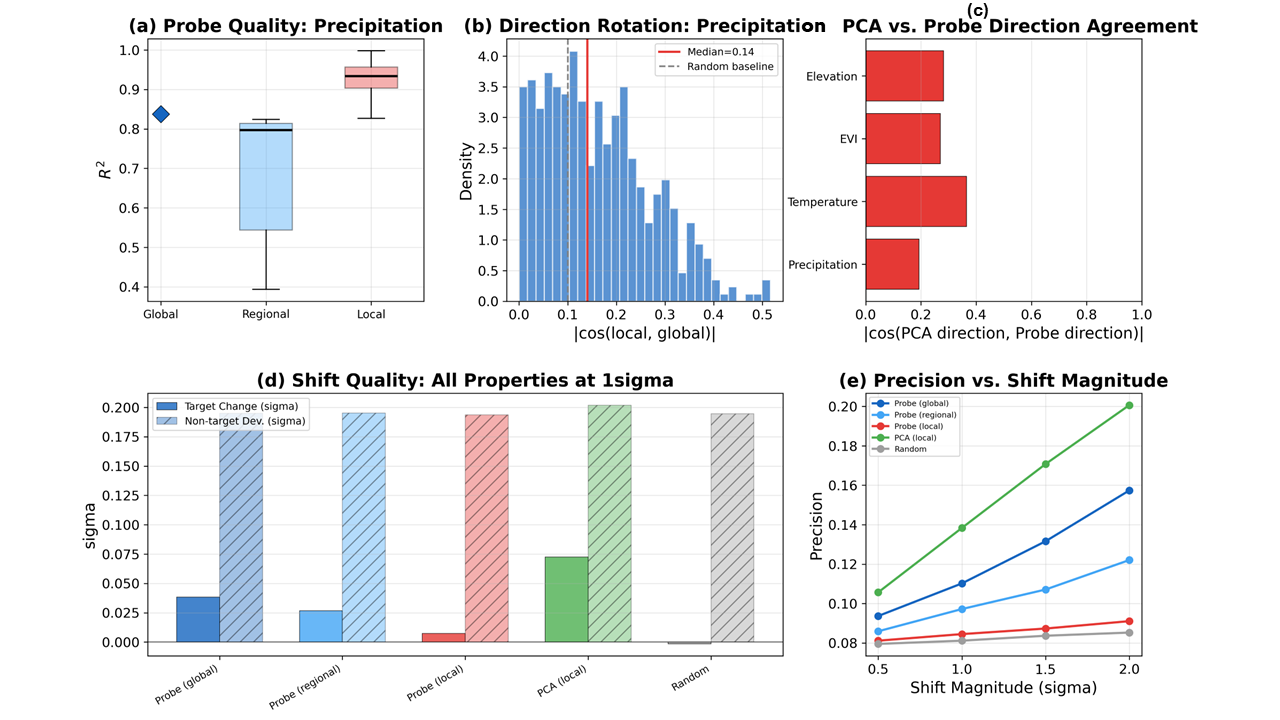}}
{\caption{Linear probes for concept directions across spatial scales. 
(a)~Probe predictive accuracy ($R^2$) at global, regional, and local 
scales for precipitation; local probes achieve the highest $R^2$ but 
with substantial variance across locations. 
(b)~Distribution of $|\cos\theta|$ between local and global probe 
directions for precipitation, with median $= 0.14$ approaching the 
random baseline of $0.10$; concept directions rotate substantially across 
the manifold. 
(c)~Agreement between PCA-derived and probe-derived global concept 
directions ($|\cos\theta|$); low values confirm that the single 
best-correlated PC is a poor proxy for the supervised concept direction. 
(d)~Targeted shift quality at $1\sigma$: mean target change (solid bars) 
and non-target deviation (hatched bars) for five methods averaged across 
all properties; PCA-derived local directions produce the largest target 
changes despite lower predictive accuracy. 
(e)~Shift precision as a function of shift magnitude; no method achieves 
high precision at any magnitude, consistent with manifold curvature 
preventing reliable compositional operations}
\label{fig:linear_probes}}
\end{figure}

However, the directions these probes identify are not spatially consistent (Figure~\ref{fig:linear_probes}a,b). The ``precipitation direction'' at one location has low commonality with the ``precipitation direction'' at another: local-global cosine similarity is $0.14$ for precipitation, $0.21$ for temperature, $0.28$ for EVI, and $0.36$ for elevation, against a random baseline of $0.10$ in 64 dimensions. Even elevation, the most geometrically stable property, shows substantial rotation. This is independent confirmation of the manifold curvature documented in RQ1: the same geometric heterogeneity that causes tangent spaces to rotate also causes concept directions to rotate.

The consequence for arithmetic is direct. When we repeat the targeted shift experiment using probe-derived directions, they produce smaller target changes than the PCA-derived directions ($+0.007\sigma$ vs.\ $+0.073\sigma$ at $1\sigma$ shift; 
Figure~\ref{fig:linear_probes}c,d), despite their superior predictive accuracy. A direction optimized to explain variance among 100 local neighbors does not generalize one step beyond that neighborhood on a curved manifold. We note that these results reflect the specific combination of AlphaEarth's 64-dimensional representation, Ridge regression probing, and the shift-and-retrieve evaluation protocol; different architectures, probing strategies, or embedding dimensionalities may behave differently. An alternative interpretation is that the direction rotation simply reflects the well-known variation of Earth surface processes across space, and that local probes trained on small neighborhoods capture local covariance structure that need not generalize. Under this reading, the finding is not that the manifold is pathologically curved, but that locally trained concept directions are not the right tool for global compositional operations. Both interpretations lead to the same practical conclusion: retrieval over the existing embedding database is more reliable than algebraic construction of new embeddings, regardless of how the concept directions are derived.

\begin{figure}[t]%
\FIG{\includegraphics[width=0.9\textwidth]{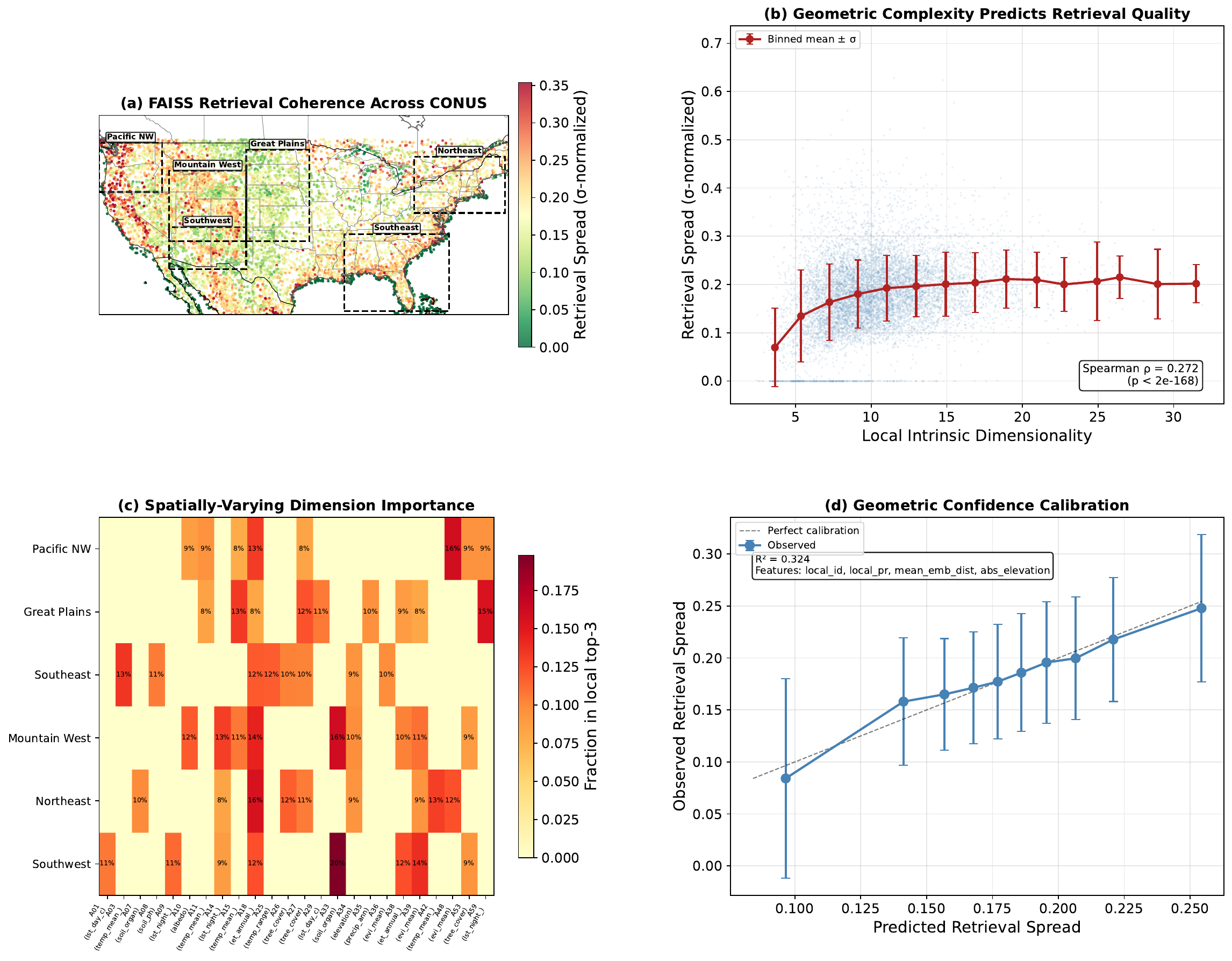}}
{\caption{Retrieval coherence and regional geometric profiles. (a) Spatial distribution of FAISS retrieval coherence across CONUS, measured as normalized spread of environmental variables among $k = 10$ nearest neighbors; warmer colors indicate less coherent retrieval. Regional bounding boxes delineate the six subregions. (b) Local intrinsic dimensionality versus retrieval spread at 10{,}000 probe locations, with binned means and standard deviations; Spearman $\rho = +0.272$ indicates that more geometrically complex neighborhoods produce less coherent retrieval. (c) Spatially varying dimension importance: each cell shows the fraction of local top-3 importance held by a given dimension within each of the six CONUS subregions, with dimension A18 (evapotranspiration) appearing among the top-ranked dimensions in five of the six regions. (d) Confidence model calibration: observed versus predicted retrieval spread using five geometric features ($R^2 = 0.32$), with the identity line shown for reference}
\label{fig:retrieval_coherence}}
\end{figure}

\paragraph{Retrieval coherence.}
While the arithmetic experiments yield low precision, retrieval produces physically coherent results across most of the manifold (Figure~\ref{fig:retrieval_coherence}a). A linear model predicting retrieval coherence from five geometric features achieves $R^2 = 0.32$ (Figure~\ref{fig:retrieval_coherence}b), with mean embedding distance to neighbors as the strongest predictor ($\rho = +0.484$), followed by local intrinsic dimensionality ($\rho = +0.272$). This means the agent system can estimate retrieval reliability at any location using only geometric features, without consulting ground-truth environmental data.

\paragraph{Regional profiles.}
Retrieval coherence varies substantially across CONUS subregions (Figure~\ref{fig:retrieval_coherence}c). The Great Plains exhibits the highest coherence (mean spread = 0.144), consistent with its low intrinsic dimensionality and environmentally homogeneous landscape. The Pacific Northwest shows the lowest coherence (0.218), where steep environmental gradients in precipitation and vegetation create more heterogeneous neighborhoods. The dominant dimensions differ across regions: temperature dimensions (A59, A15, A27) dominate the Great Plains, vegetation dimensions (A48, A18, A11) dominate the Pacific Northwest, and soil, hydrology, and terrain-linked dimensions (A33, A18, A14) dominate the Mountain West. Dimension A18 (evapotranspiration) appears among the top locally important dimensions in five of six regions (Figure~\ref{fig:retrieval_coherence}c), making it the most spatially universal dimension.

\begin{figure}[t]%
\FIG{\includegraphics[width=0.9\textwidth]{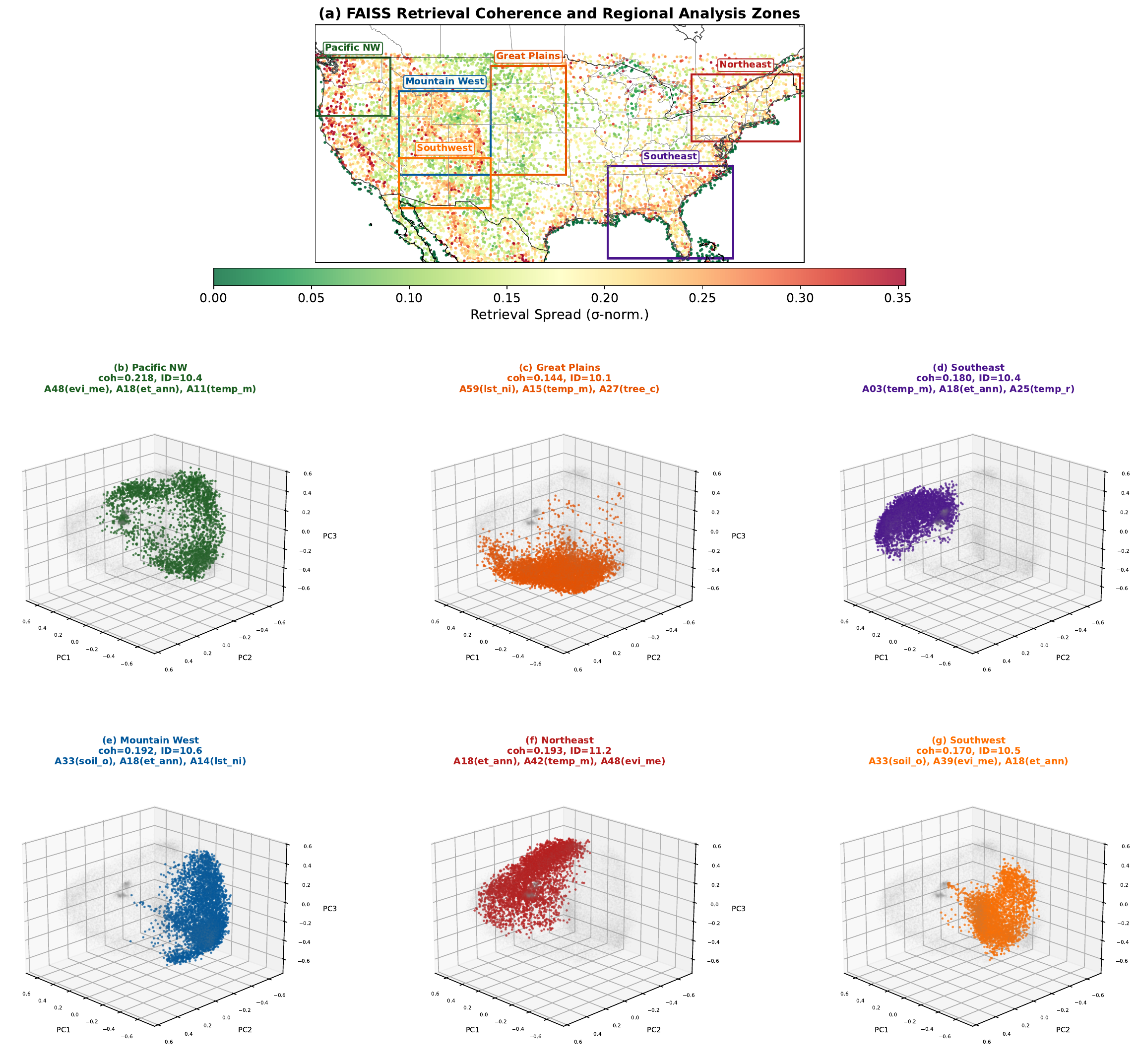}}
{\caption{Regional structure of the AlphaEarth embedding manifold. (a) FAISS retrieval coherence across CONUS (normalized spread of environmental variables among $k = 10$ nearest neighbors) with the six analysis subregions overlaid. (b--g) Three-dimensional PCA projections of the embedding space, with points belonging to each subregion highlighted in color and the remaining CONUS samples shown in gray. Each panel is annotated with the region's mean retrieval coherence (coh), mean local intrinsic dimensionality (ID), and top three locally important dimensions. Subregions occupy visibly distinct neighborhoods of the manifold, and the dominant dimensions differ across regions, reinforcing that physically meaningful regional signal is carried by the relationships among groups of dimensions rather than by any single dimension in isolation}
\label{fig:regional_3d}}
\end{figure}

To elaborate, labeling A59 a ``temperature dimension'' does not mean that A59 alone encodes temperature in a disentangled sense. \citet{rahman2026physically} showed that environmental encodings are distributed across multiple correlated embedding dimensions, and the participation ratio of 13.3 shows that variance is spread across many axes rather than concentrated in isolated ones. When we say temperature dimensions dominate a region, we mean that the local principal components with the highest loadings on those dimensions explain the most local variance, not that any single dimension is a clean proxy for temperature. The physically meaningful signal is carried by the geometric relationships among groups of dimensions, and those relationships change across the manifold, which is precisely why the regional profiles differ. Figure~\ref{fig:regional_3d} visualizes this directly: each subregion occupies a distinct neighborhood of the embedding manifold, and the dominant dimensions annotated on each panel are the ones whose joint loadings define that neighborhood.

\subsection{RQ3: Can an Agentic System Extend Satellite-Grounded Intelligence to Multi-Step Reasoning?}

\begin{figure}[t]%
\FIG{\includegraphics[width=0.9\textwidth]{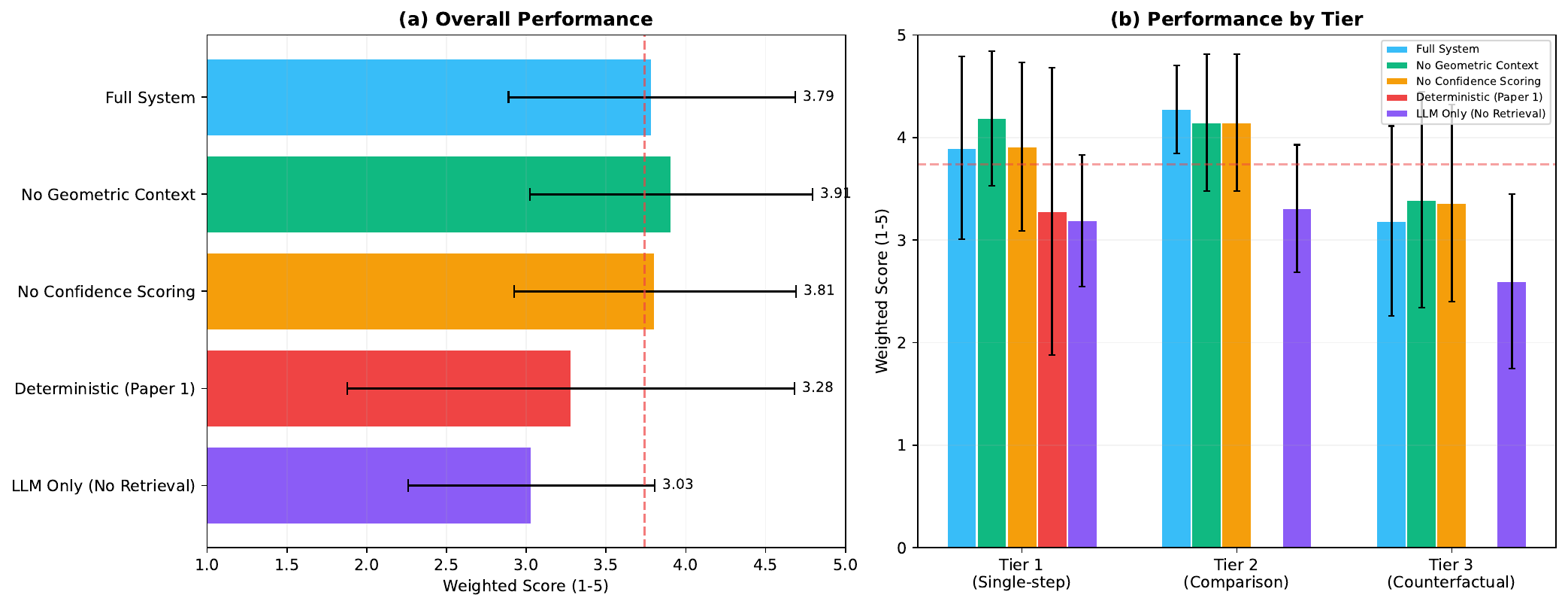}}
{\caption{Ablation study results (Claude Sonnet~4.5 system, Gemma-3-27B judge). (a) Overall weighted scores by ablation condition, with the \citet{rahman2026physically} - (Paper 1) baseline ($\mu = 3.74$) shown as a dashed reference line. Error bars indicate one standard deviation. (b) Weighted scores by query tier and condition. Tier~2 (multi-step comparison) produces the highest scores across all agentic conditions; the deterministic pipeline can only process Tier~1 queries}
\label{fig:ablation}}
\end{figure}

\paragraph{Retrieval grounding is the dominant factor.}
The largest effect in the ablation study is the difference between satellite-grounded and parametric-only generation (Table~\ref{tab:ablation_results}, Figure~\ref{fig:ablation}a). The full system achieves $\mu = 3.79 \pm 0.90$ versus $\mu = 3.03 \pm 0.77$ for the LLM-only baseline ($\Delta = +0.76$). The grounding criterion shows the most significant contrast: $G = 3.78$ for the full system versus $G = 1.65$ for LLM-only, indicating that without FAISS retrieval, the language model lacks access to location-specific environmental data and produces ungrounded responses.

\paragraph{The agentic architecture enables new query classes.}
The agent achieves its highest scores on Tier~2 (multi-step comparison) queries: $\mu = 4.28 \pm 0.43$, the highest score and lowest variance of any tier--condition combination (Table~\ref{tab:per_tier}, Figure~\ref{fig:ablation}b). These queries require the agent to resolve multiple locations, retrieve embeddings for each, and synthesize a comparative assessment. The deterministic pipeline from \citet{rahman2026physically} cannot decompose such queries at all; it achieves $\mu = 3.28 \pm 1.40$ on Tier~1 queries only. The agent's ability to plan multi-step tool-call sequences is essential for this query class.

\paragraph{Geometric context provides manifold-aware grounding.}
With Claude Sonnet~4.5 as the system model, removing geometric tools slightly improves the weighted score ($\mu = 3.91$ vs.\ $\mu = 3.79$; $\Delta = -0.12$). This occurs because geometric tools increase planning complexity (mean 8.8 tool calls with geometric context vs.\ 6.0 without), and the additional rounds introduce more opportunities for planning errors. However, the geometric grounding score tells a different story: the full system achieves $\text{GG} = 2.64$ versus $\text{GG} = 2.10$ without geometric tools, indicating that the system does reference manifold properties more frequently when geometric tools are available. The standard evaluation criteria, designed for \citet{rahman2026physically}'s single-step pipeline, do not reward this form of uncertainty-aware reasoning.

\begin{table}[t]
\tabcolsep=0pt%
\TBL{\caption{Ablation results (Claude Sonnet~4.5 system, Gemma-3-27B judge). G = grounding, A = scientific accuracy, C = completeness, H = coherence, U = practical utility, GG = geometric grounding (Tier~2--3 only, not included in weighted score). \citet{rahman2026physically} baseline: $\mu = 3.74 \pm 0.77$}\label{tab:ablation_results}}
{\begin{fntable}
\begin{tabular*}{\textwidth}{@{\extracolsep{\fill}}lcccccccc@{}}\toprule
\TCH{Condition} & \TCH{$n$} & \TCH{$\mu \pm \sigma$} & \TCH{G} & \TCH{A} & \TCH{C} & \TCH{H} & \TCH{U} & \TCH{GG} \\\midrule
Full system & 120 & $3.79 \pm 0.90$ & 3.78 & 3.81 & 3.43 & 4.61 & 3.41 & 2.64 \\
No geometric & 120 & $3.91 \pm 0.89$ & 3.96 & 3.91 & 3.54 & 4.70 & 3.52 & 2.10 \\
No confidence & 120 & $3.81 \pm 0.88$ & 3.81 & 3.79 & 3.43 & 4.64 & 3.48 & 2.91 \\
Deterministic & 40 & $3.28 \pm 1.40$ & 3.02 & 3.62 & 2.58 & 4.38 & 2.98 & --- \\
LLM only & 120 & $3.03 \pm 0.77$ & 1.65 & 3.64 & 2.92 & 4.65 & 2.83 & 1.05 \\
\botrule
\end{tabular*}
\end{fntable}}
\end{table}

\begin{table}[t]
\tabcolsep=0pt%
\TBL{\caption{Per-tier weighted scores ($\mu \pm \sigma$). The deterministic pipeline can only process Tier~1 queries. Tier~2 queries consistently produce the highest scores across agentic conditions}\label{tab:per_tier}}
{\begin{fntable}
\begin{tabular*}{\textwidth}{@{\extracolsep{\fill}}lccc@{}}\toprule
\TCH{Condition} & \TCH{Tier~1} & \TCH{Tier~2} & \TCH{Tier~3} \\\midrule
Full system & $3.90 \pm 0.89$ & $4.28 \pm 0.43$ & $3.19 \pm 0.93$ \\
No geometric & $4.19 \pm 0.66$ & $4.15 \pm 0.67$ & $3.39 \pm 1.05$ \\
No confidence & $3.91 \pm 0.82$ & $4.15 \pm 0.67$ & $3.36 \pm 0.96$ \\
Deterministic & $3.28 \pm 1.40$ & --- & --- \\
LLM only & $3.19 \pm 0.64$ & $3.31 \pm 0.62$ & $2.60 \pm 0.85$ \\
\botrule
\end{tabular*}
\end{fntable}}
\end{table}

\subsection{RQ4: Does Geometric Metadata Utility Depend on Model Reasoning Capability?}

The cross-model benchmark comparing Claude Sonnet~4.5 and Opus~4.6 on 36 queries reveals that the direction of the geometric context effect reverses between models (Table~\ref{tab:model_scaling}). Sonnet values reported in this comparison are the full (n=120) ablation results from Table~\ref{tab:ablation_results}; Opus was evaluated on a (n=36) subset (12 per tier) from the same query set. The Opus aggregate and per-tier means are therefore noisier than the Sonnet reference values, which we account for in the interpretation below. With Sonnet, geometric context reduces the weighted score ($\Delta = -0.12$); with Opus, it improves the score ($\Delta = +0.07$). The aggregate effect with Opus is small, but the pattern is most pronounced at Tier~1, where Opus achieves $\Delta = +0.58$ (full: $4.01 \pm 0.95$, no geometric: $3.43 \pm 1.19$), while Sonnet shows the opposite direction at Tier~1 ($\Delta = -0.29$). This suggests that confidence calibration and regional profiles improve single-location assessments when the model can reason about them effectively, but impose a planning burden on less capable models.

The geometric grounding scores provide further evidence. Opus references manifold properties substantially more than Sonnet when geometric tools are available: $\text{GG} = 3.38$ (Opus) vs.\ $\text{GG} = 2.64$ (Sonnet), a 28\% increase. Even without geometric tools, Opus achieves higher GG ($2.50$ vs.\ $2.10$), suggesting that the stronger model draws on geometric concepts from its parametric knowledge more readily. At the same time, Opus uses more tool calls on average ($12.6$ vs.\ $8.8$ for Sonnet in the full condition) and produces zero short responses ($< 300$ characters), indicating that the stronger model absorbs the planning complexity of geometric tools without the failure modes observed with Sonnet.

We note that the aggregate $\Delta = +0.07$ for Opus is within the noise expected for $n = 36$ and cannot be interpreted as a statistically robust effect on the weighted score alone. The more reliable evidence for the sign flip comes from the per-tier breakdown (where Tier~1 $\Delta = +0.58$ is large relative to variation) and from the geometric grounding scores (which show a consistent and substantial advantage for Opus across conditions). Together, these point toward a relationship between model capability and geometric metadata utility, though confirming this pattern will require evaluation across a broader range of models and larger query sets.

\begin{table}[t]
\tabcolsep=0pt%
\TBL{\caption{Cross-model benchmark: geometric context contribution by model. Sonnet values (from the full $n = 120$ ablation in Table~\ref{tab:ablation_results}) are reproduced here to anchor the Opus comparison; Opus was evaluated on a $n = 36$ query subset (12 per tier) drawn from the same set. $\Delta = \mu_{\text{full}} - \mu_{\text{no\_geo}}$. GG = geometric grounding, scored on Tier~2--3 only}\label{tab:model_scaling}}
{\begin{fntable}
\begin{tabular*}{\textwidth}{@{\extracolsep{\fill}}lccccc@{}}\toprule
\TCH{Model} & \TCH{Full ($\mu \pm \sigma$)} & \TCH{No geo ($\mu \pm \sigma$)} & \TCH{$\Delta$} & \TCH{GG (full)} & \TCH{GG (no geo)} \\\midrule
Sonnet~4.5 & $3.79 \pm 0.90$ & $3.91 \pm 0.89$ & $-0.12$ & 2.64 & 2.10 \\
Opus~4.6 & $3.85 \pm 0.91$ & $3.78 \pm 0.94$ & $+0.07$ & 3.38 & 2.50 \\
\botrule
\end{tabular*}
\end{fntable}}
\end{table}

\section{Conclusion} \label{sec:conclusion}

We have characterized the manifold geometry of Google AlphaEarth foundation model embeddings across 12.1 million samples in the Continental United States and developed an agentic system that leverages this geometric understanding for environmental reasoning. The embedding space occupies approximately 10--13 effective dimensions within the 64-dimensional ambient space, is genuinely curved (84\% of locations exhibit tangent angles above 60\textdegree), and exhibits local geometry that differs substantially from global structure (local--global alignment of 0.17 versus a random baseline of 0.125). This curvature means that the directions associated with a given environmental property, whether identified by PCA or by supervised linear probing, rotate from one location to another rather than remaining globally consistent, motivating retrieval-based reasoning as the primary operational strategy.

The agentic system built on these findings demonstrates three results. First, satellite embedding retrieval is the dominant contributor to response quality, accounting for a 0.76-point improvement over parametric-only generation. Second, the agentic architecture enables multi-step reasoning over the embedding database, achieving its strongest performance on comparison queries that deterministic pipelines cannot process. Third, the utility of geometric metadata varies with the reasoning capability of the consuming language model: a stronger model (Opus~4.6) references geometric properties 28\% more frequently than a weaker model (Sonnet~4.5) and shows positive returns from geometric context at the per-tier level, while the weaker model incurs a net cost from the additional planning complexity. This pattern suggests that as language models improve, the value of detailed geometric characterization of embedding spaces may grow, though confirming this will require evaluation across a wider range of models.

The geometric dictionary and agentic architecture developed here are not specific to AlphaEarth. Any dense embedding space with curved geometry and spatially varying structure could benefit from analogous characterization and geometry-aware tool design. We release our analysis code and the enhanced geometric dictionary to support further work on interpretable, geometry-informed geospatial intelligence systems.

\begin{Backmatter}

\paragraph{Acknowledgments}
The authors thank Dartmouth Libraries for their support during this research. The agentic system and all LLM-based evaluation were built using the Dartmouth Chat API (chat.dartmouth.edu), which provided access to multiple large language model backends including Claude and Gemma.

\paragraph{Funding Statement}
None.

\paragraph{Competing Interests}
S.J.B. is employed by LGND AI, and C.L. is the Director of TipplyAI. The research was conducted independently of any commercial activities of either company. M.R. declares no competing interests.

\paragraph{Data Availability Statement}
The manifold characterization code (Phases 1-2) is publicly available at \url{https://github.com/mashrekur/alpha_geom} (archived at https://doi.org/10.5281/zenodo.19652729). The repository includes representative evaluation responses with judge scores, stratified by complexity tier, ablation condition, and system model. All input datasets used in this study are publicly available through Google Earth Engine. The extracted dataset of co-located AlphaEarth embeddings and environmental variables is the same as described in \citet{rahman2026physically} and is available from the corresponding author upon reasonable request. The agentic system and evaluation pipeline depend on the Dartmouth Chat API \citep{dartmouth_chat}, which requires institutional authentication credentials issued by Dartmouth College; these components are available from the corresponding author upon request.

\paragraph{Ethical Standards}
The research meets all ethical guidelines, including adherence to the legal requirements of the study country.

\paragraph{Author Contributions}
Conceptualization: MR, SJB, CL Methodology: MR, SJB, CL  Data curation: MR Data visualisation: MR Writing original draft: MR Writing review and editing: MR, SJB, CL. 

\printbibliography

\end{Backmatter}

\end{document}